%% file: neurips_2023.tex
\documentclass{article}

\usepackage[final]{neurips_2023}

\usepackage[utf8]{inputenc} 
\usepackage[T1]{fontenc}    
\usepackage{hyperref}       
\usepackage{url}            
\usepackage{booktabs}       
\usepackage{amsfonts}       
\usepackage{nicefrac}       
\usepackage{microtype}      

\usepackage{graphicx}
\usepackage{subfigure}
\usepackage[table,x11names]{xcolor}

\usepackage{amsmath}
\usepackage{amssymb}
\usepackage{mathtools}
\usepackage{multirow}
\usepackage{amsthm}

\definecolor{lightgreen}{HTML}{CCFF99}
\definecolor{best}{HTML}{a4dbb6}
\definecolor{light}{HTML}{f2c0c0}
\definecolor{darkred}{HTML}{941e1e}
\definecolor{background}{HTML}{ccff99} 
\definecolor{highlight}{HTML}{33cc33} 

\theoremstyle{plain}
\newtheorem{theorem}{Theorem}[section]

\theoremstyle{definition}

\theoremstyle{remark}

\title{Test-Time Amendment with a Coarse Classifier for Fine-Grained Classification}

\author{
Kanishk Jain$^{1}$, Shyamgopal Karthik$^{2}$, Vineet Gandhi$^{1}$\\
$^1$IIIT Hyderabad $^2$University of Tübingen 
}

\begin{document}

\maketitle

\begin{abstract}
 We investigate the problem of reducing mistake severity for fine-grained classification. Fine-grained classification can be challenging, mainly due to the requirement of domain expertise for accurate annotation. However, humans are particularly adept at performing coarse classification as it requires relatively low levels of expertise. To this end, we present a novel approach for Post-Hoc Correction called Hierarchical Ensembles (HiE) that utilizes label hierarchy to improve the performance of fine-grained classification at test-time using the coarse-grained predictions. By only requiring the parents of leaf nodes, our method significantly reduces \emph{avg. mistake severity} while improving \emph{top-1} accuracy on the \textit{iNaturalist-19} and \textit{tieredImageNet-H} datasets, achieving a new state-of-the-art on both benchmarks. We also investigate the efficacy of our approach in the semi-supervised setting. Our approach brings notable gains in \emph{top-1} accuracy while significantly decreasing the severity of mistakes as training data decreases for the fine-grained classes. The simplicity and post-hoc nature of HiE renders it practical to be used with any off-the-shelf trained model to improve its predictions further. The code is available at: \url{https://github.com/kanji95/Hierarchical-Ensembles}
\end{abstract}

\section{Introduction}
\begin{figure}[t]
    \centering
    \includegraphics[width=0.9\columnwidth]{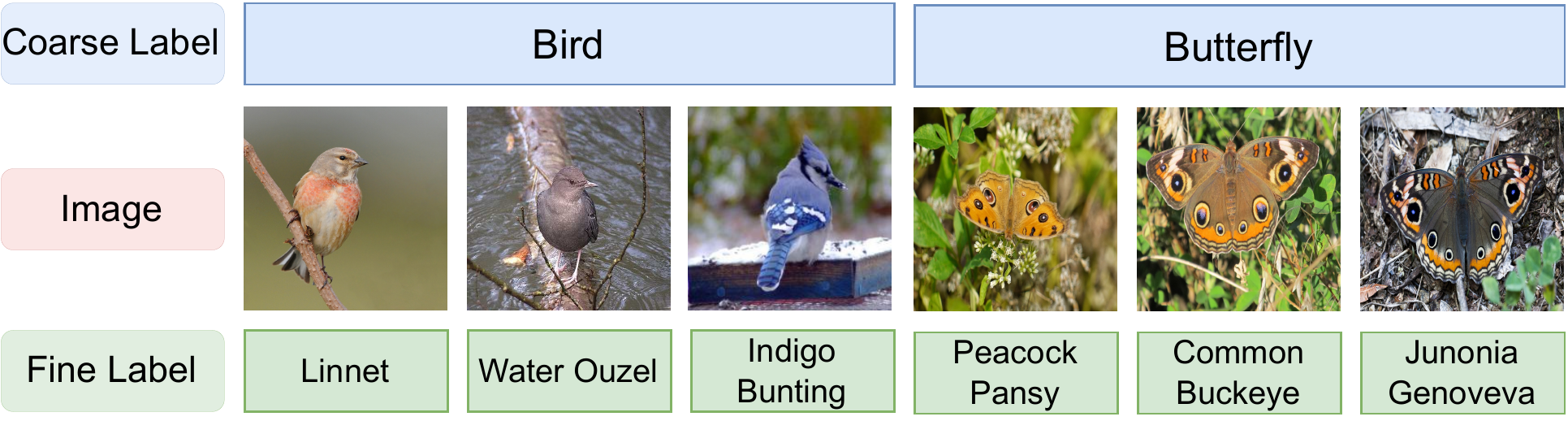}
    \caption{While it is straightforward to identify the images based on their coarse labels, it is challenging to distinguish them based on fine-grained labels without domain expertise.}
    \label{fig:teaser}
\end{figure}

Over the past decade, large-scale datasets have been instrumental in driving the rapid progress of computer vision/image recognition. However, in settings that require experts to annotate samples, collecting large amounts of labeled data can be prohibitively expensive. \emph{Fine-Grained Visual Classification (FGVC)} is one such example, where one would need a domain expert to be able to identify the category for a particular sample. However, it can be cheaper to obtain coarse labels for the same samples in these settings. For instance, in Figure~\ref{fig:teaser} while it is evident to spot and identify a butterfly or a bird, one would need a \emph{Lepidopterist} or an \emph{Ornithologists} to categorize them as \emph{Junonia Genoveva} or \emph{Water Ouzel}.



Therefore, one popular research direction in recent times has been to utilize the availability of a larger set of images with coarse labels to improve the performance (i.e \emph{top-1} accuracies) of the neural network models on various fine-grained classification benchmarks~\cite{semi-inat,su2021semi}. Using coarse labels in a classification setting introduces the concept of a label hierarchy, where all the labels in a dataset are connected through some taxonomy. This taxonomy could either be defined from the biological taxonomy, as with many species recognition datasets or could be derived from language-based ontologies such as WordNet~\cite{wordnet}. Another mainstream usage of label hierarchies in recent times is to use them to reduce the severity of mistakes committed by various classification models~\cite{deng2010does,barz2019hierarchy, mbm,crm,haf}. For instance, mistaking a car for a bus is a better mistake than mistaking a car for a lamppost. This originated from the extensive work on cost-sensitive classification~\cite{domingos1999metacost,zadrozny2001learning}. However, defining pairwise costs can be a tedious process and scales quadratically with the number of classes. Therefore, the recent works have all focused on using the label hierarchy and defining costs automatically based on graph distances (e.g. using the height of the least common ancestor between the predicted and the ground truth class as a proxy for cost). The research in this direction aims to invent methods that at least retain the accuracies of the backbone baseline models while reducing the severity of mistakes committed by it. 

While these two research directions might seem entirely different, they work with the same inputs (i.e., images, labels, and the label hierarchy). Therefore, one would ideally want a method that can use additional coarsely labelled samples to improve the model's accuracy and utilise the label hierarchy to reduce the severity of the mistakes. However, these two goals have been tackled independently. Post-hoc correction methods~\cite{deng2010does,crm}, learning hierarchy-aware features~\cite{mbm,haf,chang2021your} as well as structured embedding spaces~\cite{barz2019hierarchy,devise} have been proposed to reduce the severity of mistakes. Coarse labels have been used for self-training/pseudo-labelling~\cite{phoo2021coarsely} and with an additional objective to regularise/enhance the training~\cite{su2021semi} in the semi-supervised learning frameworks. 

In this work, we propose a simple method that is able to bring significant improvements to both of these problems. Our fundamental insight is that models trained for different granularities focus on different aspects. This is also corroborated by recent work~\cite{chang2021your,haf}, which shows that when training a model jointly, coarse-level label prediction exacerbates fine-grained feature learning. Therefore, we train two separate models for the coarse and fine-grained labels. At inference, we employ a modified decision rule, which computes the $\mathrm{argmax}$ over the normalized score of coarse and fine grained predictions. We refer to the proposed strategy as a Hierarchical Ensemble (HiE). We find that employing HiE not only reduces the severity of mistakes, but is able to improve \emph{top-1} accuracies, which are far more pronounced in the semi-supervised setting. More formally, we make the following contributions:

\begin{enumerate}
    \item We introduce a simple post-hoc mechanism called Hierarchical Ensembles for incorporating cues from multiple models trained for different granularities of a hierarchy tree.
    \item We demonstrate that HiE brings considerable reductions in mistake severity while improving the \emph{top-1 accuracy}. HiE significantly outperforms two competent baselines and six other methods from the prior art and achieves state-of-the-art performance on both \textit{iNaturalist-19} and \textit{tieredImageNet-H} benchmarks.
     \item  We illustrate that HiE brings consistent gains when plugged with existing semi-supervised methods exploiting additional coarse labeled data to improve the fine-grained classification. On \textit{iNaturalist-19} dataset, HiE combined with Moco~\cite{moco} and hierarchical loss~\cite{su2021semi}, recovers 86\% of the underlying performance while using only 0.5\% of the fine-grained annotations.  
    
\end{enumerate}

\section{Related Work}
\textbf{Learning from Taxonomic Labels:} There has been a long history of utilizing label hierarchies for better classification performance. Sila and Freitas~\cite{silla2011survey} perform a comprehensive survey of hierarchical classification methods applied to various tasks in different application domains. Broadly, the methods that train hierarchy-aware features can be categorized into three categories a) \emph{Label-Embedding methods}, b) \emph{Hierarchical architectures},  and c) \emph{Hierarchical loss functions}. 

The label-embedding methods project the categories into a semantic embedding space~\cite{devise,latem} instead of viewing them as one-hot encodings and are widely used, especially in zero-shot learning~\cite{xian2018zero}. The semantic embedding space can be derived either from side-information such as class attributes~\cite{latem}, taxonomies~\cite{mbm}, or from natural language~\cite{devise}. These can be further enhanced by learning these embeddings on a hypersphere~\cite{barz2019hierarchy} or hyperbolic spaces~\cite{nickel2017poincare,khrulkov2020hyperbolic,liu2020hyperbolic}. 

On the other hand, hierarchical loss functions try to modify the training objective to incorporate the label hierarchy~\cite{wu2016learning,mbm,garnot2020leveraging,haf,valmadre2022hierarchical}. The most common approach here is to introduce a loss at every level in the hierarchy~\cite{wu2016learning,mbm,haf}. The challenge in this direction has been to ensure that the coarse-grained task does not hamper the accuracy of the fine-grained classifier~\cite{chang2021your,haf}. 

There have also been attempts to modify the architecture based on the label hierarchy. For instance, popular ideas revolve around having branches at different layers of the model~\cite{bilal2017convolutional}, predicting the conditional probabilities at each node in the hierarchy~\cite{yolov2, ridnik2021imagenet}, or partitioning the feature space using the levels in the hierarchies~\cite{chang2021your}. 

Most closely related to our work are the works that attempt to incorporate the label hierarchy information post-hoc during inference. Here, traditional Conditional Risk Minimization (CRM)~\cite{duda1973pattern} was applied in this setting by \cite{deng2010does} and recently revisited by \cite{crm}. Our proposed approach is also applied during inference; however, it is orthogonal to CRM, and we demonstrate that CRM can be used to improve our results further. Concurrent to our work, \cite{chils} proposed a similar approach to utilize subclasses to improve superclass recognition performance in a post-hoc manner with vision-language models~\cite{clip}. 

\textbf{Fine-Grained Visual Classification:} Contrary to the popular image recognition benchmarks, fine-grained visual classification focuses on recognizing the differences between similar-looking categories~\cite{inat,birds,aircraft}. Here, methods often focus on learning discriminative features that use local information since most categories share a similar global structure~\cite{angelova2013efficient,lin2015deep,zhang2014part}. However, the most related to our work are the ones that use class taxonomies to improve fine-grained classification performance. This is most frequently done in the semi-supervised setting~\cite{semi-inat}, where one has access to an additional set of weakly labeled samples. Here, approaches have explored incorporating a hierarchical loss on the coarse labels~\cite{su2021semi} or using the coarse labels to filter pseudo-labels~\cite{phoo2021coarsely} in addition to the standard semi-supervised learning techniques of consistency regularization~\cite{fixmatch}, pseudo-labeling~\cite{pseudo,pseudo++}, and contrastive learning~\cite{moco}. Our proposed approach is complementary to these works and can be applied to all of these works to bring additional improvements.

\section{Methodology}

\subsection{Problem Formulation}

We consider the fine-grained classification task with a label hierarchy $\mathcal{H}$ of $L$ levels defined over the class labels. The leaf nodes of $\mathcal{H}$ correspond to fine-grained classes at level $L$, and the class labels at level $l < L$ are considered coarse-grained labels. The goal is to use the information from coarse annotations to improve the performance on fine labels. Formally, given a dataset $\mathcal{X} = \{(x_i, {y}_{i}^{l})\mid i = 1,2, \dots, N\}$ consisting of $N$ training images and their respective ground truth labels at level $l$, where label ${y}_{i}^{l} \in \mathcal{Y}^{l} = \{1, 2, ..., N_l\}$, and $N_l$ corresponds to the number of classes at level $l$, the task is to train a classifier $f^{L}_{\theta} : \mathcal{X} \rightarrow p(\mathcal{Y}^{L})$, on fine-grained class labels by making use of the information available through coarse-grained labels $\mathcal{Y}^{l < L}$.

\subsection{Proposed Method}

\begin{figure*}[t!]
 \centering
 \includegraphics[width=0.7\linewidth]{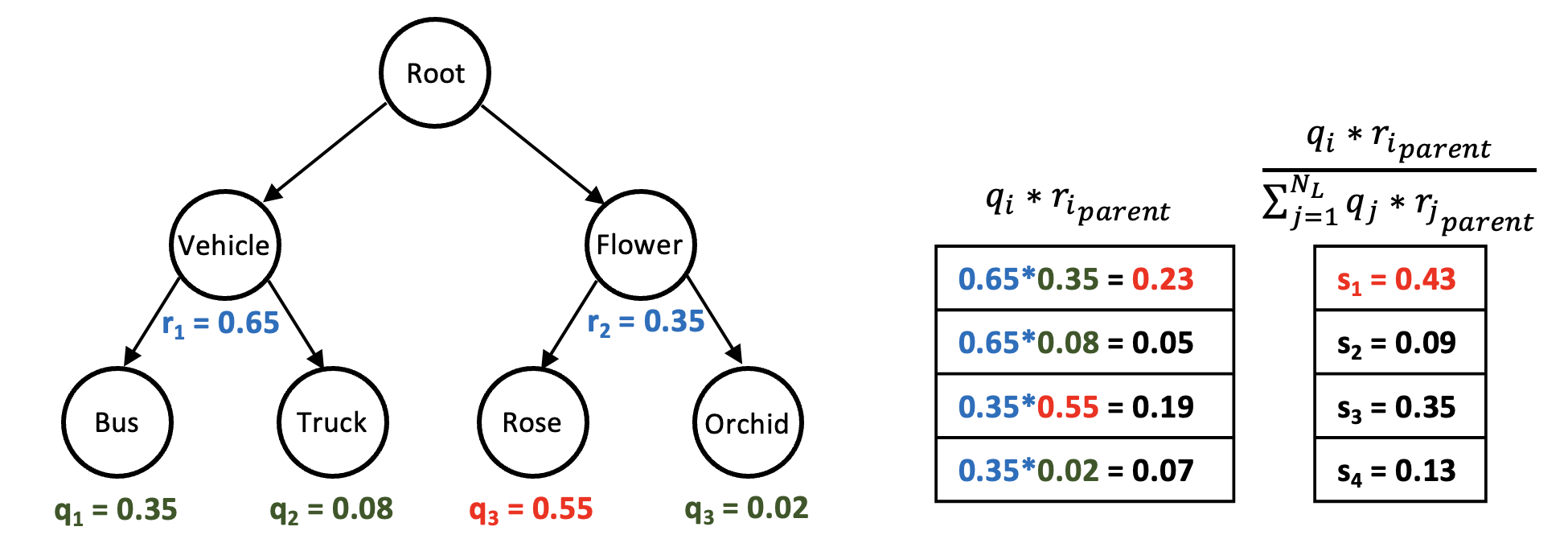} 
 \caption{Consider a four class hierarchy tree and corresponding fine-grained predictions ($q_i$) and coarse level predictions ($r_i$) obtained using independent models. Our work aims to improve the fine grained classification by leveraging the predictions at the coarse level. In the given example, the class prediction changes from `rose' to `bus' after the post-hoc correction. }
 \label{fig:method}
\end{figure*}


Suppose we have two classifiers, $f_{\theta}^{L}$ and $f_{\phi}^{L-1}$ trained on hierarchy levels $L, L-1$ with $\theta$ and $\phi$ being their parameters, respectively, then for a sample $x \sim D$ , we have, $\hat{y}_L = f_{\theta}^{L}(x)$
 and $\hat{y}_{L-1} = f_{\phi}^{L-1}(x)$ as logits of two classifiers for the input $x$. We further define:
 \begin{equation}
     Q = [q_1, q_2 , ..., q_{N_L}] = \mathrm{softmax} (\hat{y}_L) \nonumber
 \end{equation}
  \begin{equation}
     R = [r_1, r_2 , ..., r_{N_{L-1}}] = \mathrm{softmax} (\hat{y}_{L-1})  \nonumber
 \end{equation}


We reweigh the fine-grained predictions $\mathcal{Q}$, given the softmax probabilities $\mathcal{R}$ obtained from the coarse-grained classifier: 

 \begin{equation}
P(i|x, \theta, \phi, \mathcal{H}) = \frac{P(i|x; \theta) P(i_{parent}|x; \phi)}{\sum_{j=1}^{N_L} P(j|x; \theta) P(j_{parent}|x; \phi)}, \quad \text{where} ~ i_{parent} = \mathcal{H}(i)
 \end{equation}

The denominator is the normalization term, ensuring $\sum_{i=1}^{N_L} P(i | x, \mathcal{H}) =1 $. In the studied setting, $P(i | x, \mathcal{H})$ simplifies to: 
  \begin{equation}
P(i | x, \mathcal{H}) = s_i = \frac{q_i \mbox{ . } r_{i_{parent}}}{\sum_{j=1}^{N_L} q_j \mbox{ . } r_{j_{parent}}}
 \end{equation}


The main modification we make in the classification strategy at the fine grained level is in the decision rule, which now selects the class that maximizes the normalized score:
\begin{equation}
\underset{i}{\mathrm{argmax}} \mbox{ } P(i | x, \mathcal{H})
 \end{equation}


In Figure~\ref{fig:method}, we illustrate a four-class example, comparing standard cross-entropy predictions with those from our proposed method. We would like to point out that the proposed technique can be interpreted as a Hierarchical Ensemble. Existing literature in neural networks-based ensemble learning has often focused on classifiers trained at a single granularity~\cite{wang2020wisdom} and has largely overlooked ensembles with hierarchical context. In this work, we highlight the practical benefits of hierarchical ensembles for improving the fine-grained classification task by choosing the simplest possible method of multiplying the predictions at different levels of hierarchy. 

We show that if we make a correct prediction at the coarse level, the proposed Hierarchical Ensemble (HiE) is guaranteed to improve the downstream predictions at the fine-grained classification task. 
\begin{theorem}
\label{th:Posthoc1}
Assuming, $Q = [q_1, q_2 , ..., q_{N_L}]$ and  $R = [r_1, r_2 , ..., r_{N_{L-1}}]$ are the predictions obtained at the fine and coarse grained labels for a given input $x$, such that $\sum_{i=1}^{N_L} q_i = 1$ and $\sum_{i=1}^{N_{L-1}} r_i = 1$. For the ground truth labels $g$ and $g_{parent}$ at the fine grained and the coarse grained levels respectively:

Now assuming that the coarse label is correctly predicted by the coarse prediction network i.e. $\mathrm{argmax} (R) = g_{parent}$, we wish to prove that:
 \begin{equation}\nonumber
      \frac{q_g \mbox{ . } r_{g_{parent}}}{\sum_{j=1}^{N_L} q_j \mbox{ . } r_{j_{parent}}} \ge q_g.
 \end{equation}



\begin{proof} The denominator iterates over the fine grained predictions and multiplies them with their parent's prediction scores. This is equivalent to iterating over the coarse label predictions and multiplying with the sum of prediction scores for all its children. By rewriting the denominator, we obtain:\\
 \begin{equation}\nonumber
\sum_{j=1}^{N_L} q_j \mbox{ . } r_{j_{parent}} = \sum_{j=1}^{N_{L-1}} r_j \sum_{i\in j_{child}} q_{i}  
 \end{equation}
Assuming, $\sum_{i\in j_{child}} q_{i} = z_j$
 \begin{equation}\nonumber
 \sum_{j=1}^{N_{L-1}} r_j \sum_{i\in j_{child}} q_{i} = \sum_{j=1}^{N_{L-1}}  r_j \mbox{ . } z_j = R^T Z
  \end{equation}

Invoking Holder's inequality, using $a=\infty$ and $b=1$, ($1/a + 1/b = 1$), we obtain:
 \begin{equation}\nonumber
 \begin{split}
 R^T Z &\le {\|R\|}_{\infty} { \|Z\| }_{1} \\ 
 \end{split}
\end{equation}  

Since, $\|R\|_{\infty} = \mathrm{max} (R) = r_{g_{parent}}$ and $\|Z\|_{1} = \sum_{i=1}^{N_L} q_i = 1$. We can say:

\begin{equation}\nonumber
  R^T Z \le {\|R\|}_{\infty} { \|Z\| }_{1} \\ \le r_{g_{parent}}
\end{equation} 

Given the above equation, we can conclude that: 
 \begin{equation}\nonumber
      \frac{r_{g_{parent}}}{\sum_{j=1}^{N_L} q_j \mbox{ . } r_{j_{parent}}} \ge 1.
 \end{equation}
\end{proof}

\end{theorem}

When correct predictions are made at both the coarse and fine-grained levels, the method will help increase the confidence of the prediction. A more interesting case occurs when the coarse-level prediction is correct; however, the fine-grained classifier makes a wrong prediction. The hierarchical ensemble can shift the prediction to the correct sub-tree, reducing the mistake severity and potentially improving the \emph{top-1} accuracy. 

\section{Experiments \& Results}

Like prior works \cite{mbm, haf, crm}, we evaluate our approach on the \textit{iNaturalist-19} \cite{van2018inaturalist} and \textit{tieredImageNet-H} \cite{ren2018meta} datasets. The \textit{iNaturalist-19} dataset is a species classification dataset containing plants, animals, fungi, etc., and has an 8-level hierarchy. The \textit{tieredImageNet-H} dataset contains classes from a wide range of categories, including musical instruments, animal breeds, tools, and plant species. It has a 13-level hierarchy derived from the WordNet hierarchy. However, rather than using the full hierarchy of categories, for our method, we only use the leaf classes and their parent categories during training and inference. The \textit{iNaturalist-19} dataset includes 1010 fine-grained ``species'' classes and 72 coarse-grained classes for the ``genus'' taxonomy. The \textit{tieredImageNet-H} dataset has 608 leaf classes and 201 parent classes. We train classifiers for each level independently, using the same hyperparameter settings.

In addition, we also evaluate the performance of our approach in a semi-supervised setting. Specifically, we want to test the performance on fine-grained classification tasks using a large number of coarsely annotated examples and a limited number of finely annotated examples. We use the \textit{iNaturalist-19} dataset for this setting; the coarse labels for all the images are used, while for fine-grained labels, we sample a limited number of images per class label during training. We used the ``species'' and ``genus'' taxonomies for the fine and coarse level labels and performed experiments for cases when the number of images per fine label is 100, 50, 25, and 10. Similar to the supervised setting, we train separate classifiers on coarse and fine labels.

To align with prior efforts~\cite{crm, haf}, all of our experiments are performed using the ResNet-18 backbone. We randomly crop a portion of images and resize them to $224 \times 224$ resolution for both datasets. For the \textit{iNaturalist-19} dataset, we initialize the ResNet-18 weights with a pre-trained ImageNet model and train the classifiers using a customized SGD optimizer for 100 epochs, with different learning rates for the backbone network and the fully connected layer (0.01 and 0.1, respectively). For the \textit{tieredImageNet-H} dataset, we train the ResNet-18 from scratch (since it is derived from the ImageNet dataset) and use the AMSGrad variant of the Adam optimizer with a learning rate of $1e^{-5}$ for 120 epochs. We conduct five runs for each experiment in tables \ref{tab:inat} and \ref{tab:tiereimagenet} and report the mean and standard deviations.

\textbf{Evaluation Metrics: } Like previous works \cite{mbm, crm}, we validate our approach using the \emph{top-1} accuracy, \emph{avg. mistake severity} and \emph{hierarchical distance@k} metrics. \emph{Avg. mistake severity} metric computes the average height of the lowest common ancestor (LCA) between the predicted class and the ground truth class for the incorrectly classified samples. Since this metric considers only the incorrect predictions, it is possible to reduce the mistake severity by making more low cost mistakes \cite{crm}, as a consequence of which the \emph{top-1} accuracy decreases. To overcome this issue, \emph{hierarchical distance@k} is used, which measures the average height of \emph{top-k} predictions with the ground truth class for all samples. For $k>1$, this metric captures the quality of ranking imposed by the network in the label space.

\subsection{Making better mistakes}

In this section, we demonstrate that the proposed hierarchical ensembles significantly reduce the severity of mistakes compared to the current state-of-the-art methods. We compare our method with \cite{barz2019hierarchy}, YOLO-v2~\cite{yolov2}; both approaches proposed by~\cite{mbm} with varying parameters - soft-labels and HXE. We also compare against the method proposed by \cite{chang2021your} for multi-granular classification and HAF~\cite{haf} imposing consistency on prediction heads across label hierarchy. We contrast the performance of the methods mentioned above to the cross-entropy baseline.

 \input{./tables/inat.tex}

We also propose two custom baselines called Cross-Entropy-H and HiE-Self. Cross-Entropy-H obtains coarse class predictions by marginalizing the prediction of its fine-grained children classes. It then employs an additional hierarchical loss term~\cite{su2021semi} over the obtained coarse predictions. HiE-Self applies a hierarchical ensemble using the coarse predictions obtained post marginalization over fine-grained classes (instead of training a separate network for the coarse level). 

We also experiment with the CRM technique proposed by \cite{crm}. Since CRM is a post-hoc approach that reweighs the probability distribution of samples obtained from any trained model, it can be applied to pre-trained models of all the approaches. Therefore, in each of the Tables~\ref{tab:inat}-\ref{tab:tiereimagenet}, we group the results to report evaluation metrics with and without using CRM at test time.

\input{./tables/tiered-imagenet.tex}

Apart from HAF, all the other methods from the prior art trade-off the \emph{top-1} accuracy with the mistake severity (Tables~\ref{tab:inat}\&\ref{tab:tiereimagenet}). The results contrast with the problem's main objective, i.e., improve the hierarchical metrics by maintaining or improving the \emph{top-1} error. In some of the instances (e.g. HXE $\alpha = 0.6$, Soft-labels $\beta = 30$), the \emph{avg. mistake severity} is reduced by making additional low-severity mistakes (indicated by an increase in \emph{top-1} error and \emph{Hier dist@1}). The multi-head approach by \cite{chang2021your} brings minor gains, and HAF further improves upon it by introducing consistency loss among different head predictions.  

The proposed minimalist approach of Hierarchical Ensembles (HiE) significantly outperforms all the methods; on both datasets, with and without CRM. On the three metrics of \emph{top-1} error, \emph{avg. mistake severity}, and \emph{Hier dist@1}, the performance achieved by \textit{HiE without CRM} betters all other methods post-CRM. The gains achieved by HiE on \emph{Hier dist@5} and \emph{Hier dist@20} metrics are more pronounced, for instance, improving over HAF without CRM on \textit{iNaturalist-19} dataset by 17\% and 14\% respectively. These two ranking metrics compare the ordering of the classes provided by each of these classifiers. The results illustrate that HiE is able to reliably align the predictions with the hierarchy. The results are interesting given that HiE only employs the bottom two levels of the hierarchy, in contrast to other methods which utilize the entire hierarchy tree.

Compared with the baselines, on the \textit{iNaturalist-19} dataset, the Cross-Entropy-H brings minor gains on \emph{Hier dist@5} and \emph{Hier dist@20} metrics; however, the \emph{top-1} error increases by over a percent. On \textit{tieredImageNet-H}, Cross-Entropy-H fails to give any improvements. In contrast, the HiE-Self baseline is able to retain the original \emph{top-1} performance while bringing noticeable gains on \emph{Hier dist@k} metrics. It illustrates that hierarchical ensembles are helpful even when applied on a single network. However, training separate networks for coarse and fine-grained levels allows us to learn complementary features which are then combined using our HiE approach.

 
\subsection{Semi Supervised Learning}
We additionally validate our approach on fine-grained classification in the semi-supervised setting. We assume that coarse annotations are available for all the samples; however, fine-grained annotations are available only for a few samples in each class. The studied setting differs from cases~\cite{phoo2021coarsely}, which divide the fine-grained classes into two categories: base classes with abundant annotated samples and novel classes where only a few annotated samples are available. This framework~\cite {phoo2021coarsely} also limits the evaluation only to the novel classes. Our setting imposes firm limits on difficult-to-obtain fine-grained labels. In the extreme setting, we assume only ten annotated training samples are available for each fine-grained class. We also evaluate the performance across all the fine-grained classes together. 

\input{./tables/ssl.tex}

We train a cross-entropy baseline exclusively using the annotated fine-grained samples. We compare its performance with representative semi-supervised methods, including Pseudo-Label~\cite{pseudo}, Self-Training (ST) with distillation~\cite{hinton2015distilling}, Self-Supervised learning (MoCo) with distillation~\cite{moco} and a combination of Self-Supervised learning and Self Training (MoCo+ST). We also include a variation of all these methods (Pseudo-Label-H, MoCo-H, ST-H, MoCoST-H) with hierarchical supervised loss as described in \cite{su2021semi} with ``genus'' supervision. The hierarchical supervised loss is applied to the coarse predictions obtained by marginalizing the fine-grained class predictions. The proposed hierarchical ensemble is a complementary approach, and we compare the performance of the above methods with and without applying the HiE. 

The results are illustrated in Table~\ref{tab:ssl}. As expected, the performance of the cross-entropy baseline steeply drops with the reduction in training data. The performance of semi-supervised methods (Pseudo-Label, ST, MoCo, MoCo-ST) improves over the cross-entropy baseline; however, they fail to retain the overall performance. The semi-supervised methods become more effective with the reduced number of labeled examples (providing better proportional gains over the cross-entropy baseline). Incorporating hierarchical supervised loss brings remarkable gains in \emph{top-1} accuracy across all experiments. For instance, in the lowest data regime (using only ten fine-grained annotated samples for training), hierarchical supervised loss brings twofold improvement over the cross-entropy baseline. The performance gains showcase the efficacy of utilizing coarse taxonomic labels in the studied setting.

Employing the proposed Hierarchical Ensemble brings consistent gains across all the methods on the three studied evaluation metrics. The combined approach of Moco-H/PseudoLabel-H with HiE achieves a new state of the art in all settings. Using ten annotated examples for each fine-grained class, Moco-H with HiE achieves a \emph{top-1} accuracy of 55.13, which is higher than the performance of cross entropy baselines using a hundred annotated samples per fine-grained class. The performance gains of HiE are more evident on the \emph{avg. mistake severity} and \emph{Hier dist}@1 metrics, bringing profound gains over just using the hierarchical supervised loss. For instance, in 100 sample experiments, the \emph{Hier dist}@1 drops from 1.17 to 1.10 (6\% improvement) by applying hierarchical supervised loss. In contrast, applying HiE brings 24\% improvement, reducing \emph{Hier dist}@1 to 0.89.

Finally, we observe that in Table~\ref{tab:inat}, a cross-entropy baseline with Resnet-18 backbone trained on the entirety of the \textit{iNaturalist-19} dataset, i.e., 187385 examples with fine-grained labels achieves a \emph{top-1} accuracy of 63.56\%. In the semi-supervised setting using coarse-grained labels for entire data and just using 10 examples with fine-grained annotations for each class (10100 training samples), Moco-H with HiE is able to recover the significant portion of the underlying performance (achieving \emph{top-1} accuracy of 55.13), reducing the requirement of expert annotations to only 0.5\% of the total number of samples.   

\subsection{Performance across different architectures}
We also showcase the generalizability of the proposed HiE approach across different vision architectures in Table \ref{tab:arch_ablation}. We utilize commonly used vision architectures like MobileNet V3 \cite{howard2017mobilenets}, Resnet variants \cite{he2016deep}, EfficientNet \cite{tan2019efficientnet}, DenseNet121 \cite{huang2017densely}, and tiny variants of DeiT \cite{touvron2021training}, ViT \cite{dosovitskiy2020vit} and SwinT \cite{liu2021Swin}. All networks are trained with the same hyper-parameter setting. We compare the proposed HiE against the vanilla and CRM-based post-hoc approach. HiE outperforms other approaches by significant margins across different architectures and on all metrics. Furthermore, similar performance gains are observed irrespective of model capacity, avg. gain of 0.9\% in top-1 accuracy across resnet variants.

\input{./tables/architecture_ablation}
\input{./tables/multi_level.tex}

\subsection{Incorporating multiple hierarchical levels}
For all the above experiments, we only use two levels of hierarchies. As an additional ablation study, we validate the effectiveness of HiE in the presence of multiple hierarchical levels in Table \ref{tab:multi_level}. Specifically, we train a separate classifier for each hierarchical level and cascade the predictions top-down. We observe that incorporating the hierarchical information from two levels (levels 5, 6) above the leaf level gives the best performance; the gains taper off gradually as we incorporate information from higher levels. The reason for this dip in performance is that the higher levels do not have much additional information as they cover too many classes (e.g., on iNaturalist: kingdom (level 1), phylum (level 2), and class (level 3) have 3, 4 and 9 classes respectively) which does not help much in improving the fine-grained predictions further.


\section{Limitation}
The main limitation of HiE is the requirement of training an additional network for the coarse level. However, existing approaches employ a full hierarchy with a separate classification head for each hierarchical level and additional losses to enforce the hierarchical structure, making their training process complex and difficult to converge. Instead, our approach trained with vanilla cross-entropy loss works with partial hierarchy and trade-offs the extra compute for adaptability, reproducibility and simplicity in training. A future avenue for exploration would be to learn disentangled coarse and fine-grained features imitating HiE in a unified architecture. Another limitation is the assumption regarding the availability of underlying label hierarchy. To overcome this limitation, we can exploit large-language models \cite{brown2020language} to obtain label hierarchy using raw class labels.

\section{Conclusion}
We proposed using Hierarchical Ensembles (HiE) of independently trained networks over coarse and fine-grained levels of the label hierarchy. In terms of mistake severity, our proposed post-hoc correction consistently outperforms state-of-the-art methods in deep hierarchy-aware image classification by large margins in terms of decrease in \emph{avg. mistake severity} and \emph{hierarchical distance@k}, while simultaneously improving the \emph{top-1} accuracy. In the semi-supervised paradigm, we show that HiE delivers consistent performance gains when used in conjunction with off-the-shelf semi-supervised learning algorithms. We show that, comparable performance to a fully supervised baseline can be attained, even using merely 10 annotations for each fine-grained class on a large fine-grained image classification dataset encompassing 1010 classes. 

\bibliography{neurips_2023}
\bibliographystyle{plainnat}  

\newpage




\end{document}

%% file: tables/inat.tex
\begin{table*}[t]
\scriptsize
\begin{center}
\renewcommand{\arraystretch}{1.05} 
\begin{tabular}{ l  ccccc } 
\hline
\addlinespace[0.2cm]
\multicolumn{1}{c}{\multirow{2}{*}{Method}} & Top-$1$ Error($\downarrow$) & Mistakes severity($\downarrow$) & Hier dist@1($\downarrow$) & Hier dist@5($\downarrow$) & Hier dist@20($\downarrow$) \\
\cline{2-6} 
       & \multicolumn{5}{c}{Without CRM}                          \\
\addlinespace[0.1cm]
\hline
Cross-Entropy & 36.44 $\pm$ 0.061 & 2.39 $\pm$ 0.007 & 0.87 $\pm$ 0.004 & 1.97 $\pm$ 0.002 & 3.25 $\pm$ 0.002 \\
Cross-Entropy-H & 37.63 $\pm$ 0.054 & 2.37 $\pm$ 0.005 & 0.89 $\pm$ 0.009 & 1.86 $\pm$ 0.001 & 2.96 $\pm$ 0.001 \\
HiE-Self & 36.36 $\pm$ 0.047 & 2.36 $\pm$ 0.014 & 0.86 $\pm$ 0.005 & 1.44 $\pm$ 0.003 & 2.31 $\pm$ 0.004 \\
Barz \& Denzler \cite{barz2019hierarchy} & 62.63 $\pm$ 0.278 & \cellcolor{background} 1.99 $\pm$ 0.008 & 1.24 $\pm$ 0.005 & \cellcolor{background} 1.49 $\pm$ 0.005 & \cellcolor{background} 1.97 $\pm$ 0.005 \\
YOLO-v2 \cite{yolov2} & 44.37 $\pm$ 0.106 & 2.42 $\pm$ 0.003 & 1.08 $\pm$ 0.004 & 1.90 $\pm$ 0.003 & 2.87 $\pm$ 0.010 \\
HXE $\alpha$=0.1 \cite{mbm} & 41.48 $\pm$ 0.204 & 2.41 $\pm$ 0.009 & 1.00 $\pm$ 0.006 & 1.77 $\pm$ 0.011 & 2.69 $\pm$ 0.021 \\
HXE $\alpha$=0.6 \cite{mbm} & 45.45 $\pm$ 0.014 & 2.24 $\pm$ 0.006 & 1.02 $\pm$ 0.003 & 1.70 $\pm$ 0.005 & 2.55 $\pm$ 0.005\\
Soft-labels $\beta=30$ \cite{mbm} & 41.67 $\pm$ 0.134 & 2.32 $\pm$ 0.010 & 0.97 $\pm$ 0.006 & 1.50 $\pm$ 0.006 & 2.23 $\pm$ 0.005 \\
Soft-labels $\beta=4$ \cite{mbm} & 74.70 $\pm$ 0.212 & \cellcolor{highlight} 1.82 $\pm$ 0.005 & 1.36 $\pm$ 0.004 & \cellcolor{background} 1.49 $\pm$ 0.003 & \cellcolor{highlight} 1.96 $\pm$ 0.004\\
Chang et al. \cite{chang2021your} & 37.23 $\pm$ 0.175 & 2.28 $\pm$ 0.006 & 0.85 $\pm$ 0.004 & 1.75 $\pm$ 0.005 & 3.02 $\pm$ 0.008 \\
HAF \cite{haf} & \cellcolor{background} 36.40 $\pm$ 0.092 & 2.28 $\pm$ 0.012 & \cellcolor{background} 0.83 $\pm$ 0.002 & 1.62 $\pm$ 0.002 & 2.55 $\pm$ 0.003 \\
\cellcolor{best} HiE (Ours) & \cellcolor{highlight} 35.33 $\pm$ 0.037 & 2.15 $\pm$ 0.003  & \cellcolor{highlight} 0.76 $\pm$ 0.001  & \cellcolor{highlight} 1.33 $\pm$ 0.001  &  2.19 $\pm$ 0.003 \\
\addlinespace[0.1cm]
\hline
       & \multicolumn{5}{c}{With CRM}                          \\
       \cline{2-6} 
\addlinespace[0.1cm]
Cross-Entropy \cite{crm} & 36.51 $\pm$ 0.083 & 2.33 $\pm$ 0.001 & 0.85 $\pm$ 0.002 & 1.32 $\pm$ 0.001 & 1.86 $\pm$ 0.002 \\
Cross-Entropy-H & 37.70 $\pm$ 0.038 & 2.33 $\pm$ 0.011 & 0.88 $\pm$ 0.003 & 1.33 $\pm$ 0.002 & 1.87 $\pm$ 0.004 \\
HiE-Self & \cellcolor{background}{36.45 $\pm$ 0.057} & 2.34 $\pm$ 0.019 & 0.85 $\pm$ 0.003 & 1.32 $\pm$ 0.003 & 1.86 $\pm$ 0.008 \\
Barz \& Denzler & 75.61 $\pm$ 0.098 & 4.62 $\pm$ 0.032 & 3.49 $\pm$ 0.025 & 3.66 $\pm$ 0.024 & 3.84 $\pm$ 0.018 \\
YOLO-v2 & 45.17 $\pm$ 0.046 & 2.43 $\pm$ 0.001 & 1.10 $\pm$ 0.001 & 1.50 $\pm$ 0.001 & 1.99 $\pm$ 0.002 \\
HXE $\alpha$=0.1 & 41.47 $\pm$ 0.220 & 2.38 $\pm$ 0.011 & 0.99 $\pm$ 0.008 & 1.41 $\pm$ 0.006 & 1.93 $\pm$ 0.005 \\
HXE $\alpha$=0.6 & 45.60 $\pm$ 0.017 & 2.21 $\pm$ 0.008 & 1.01 $\pm$ 0.003 & 1.40 $\pm$ 0.004 & \cellcolor{highlight}{1.40 $\pm$ 0.004} \\
Soft-labels $\beta=30$ & 41.99 $\pm$ 0.126 & 2.31 $\pm$ 0.009 & 0.97 $\pm$ 0.007 & 1.40 $\pm$ 0.005 & 1.91 $\pm$ 0.005 \\
Soft-labels $\beta=4$ & 77.34 $\pm$ 0.262 & \cellcolor{highlight}{2.06 $\pm$ 0.012} & 1.60 $\pm$ 0.007 & 1.72 $\pm$ 0.008 & 2.14 $\pm$ 0.007 \\
Chang et al. \cite{chang2021your} & 37.31 $\pm$ 0.145 & 2.24 $\pm$ 0.008 & 0.84 $\pm$ 0.002 & 1.30 $\pm$ 0.002 & 1.84 $\pm$ 0.002 \\
HAF \cite{haf} &  36.48 $\pm$ 0.095 &  2.25 $\pm$ 0.012 &  \cellcolor{background}{0.82 $\pm$ 0.003} & \cellcolor{background}{1.29 $\pm$ 0.004} &  1.84 $\pm$ 0.002 \\
\cellcolor{best} HiE (Ours) & \cellcolor{highlight}{35.42 $\pm$ 0.025}  &  \cellcolor{background}{2.14 $\pm$ 0.015} & \cellcolor{highlight}{0.75 $\pm$ 0.005} & \cellcolor{highlight}{1.23 $\pm$ 0.005} &  \cellcolor{background}{1.79 $\pm$ 0.005} \\
\addlinespace[0.1cm]
\hline
\end{tabular}
\renewcommand{\arraystretch}{1} 
\end{center}
\caption{Results comparing \emph{top-1} error(\%) and hierarchical metrics for \textit{iNaturalist-19}. Results in the \textit{Top} block are reported without using CRM \cite{crm} technique and \textit{Bottom} block are reported using CRM. Methods highlighted with {\textcolor{best}{aqua green}} are the best performing methods in top-$1$ error (\%).  We use {\textcolor{highlight}{dark green}} for the top performers and {\textcolor{background}{light green}} for the runners-up in each metric.}
\label{tab:inat}
\end{table*}

%% file: tables/tiered-imagenet.tex
\begin{table*}[h!]
\scriptsize
\begin{center}
\renewcommand{\arraystretch}{1.05} 
\begin{tabular}{ l  ccccc } 
\hline
\addlinespace[0.2cm]
\multicolumn{1}{c}{\multirow{2}{*}{Method}} & Top-$1$ error($\downarrow$) & Mistakes severity($\downarrow$) & Hier dist@1($\downarrow$) & Hier dist@5($\downarrow$) & Hier dist@20($\downarrow$) \\
\cline{2-6} 
       & \multicolumn{5}{c}{Without CRM}                          \\
\addlinespace[0.1cm]
\hline
Cross-Entropy & 30.64 $\pm$ 0.030 & 7.07 $\pm$ 0.010 & 2.17 $\pm$ 0.006 & 5.70 $\pm$ 0.003 & 7.25 $\pm$ 0.003 \\
Cross-Entropy-H & 32.87 $\pm$ 0.042 & 7.13 $\pm$ 0.031 & 2.35 $\pm$ 0.003 & 5.70 $\pm$ 0.009 & 7.14 $\pm$ 0.012 \\
HiE-Self & 30.78 $\pm$ 0.054 & 7.05 $\pm$ 0.028 & 2.19 $\pm$ 0.004 & 5.35 $\pm$ 0.009 & 6.92 $\pm$ 0.007 \\
Barz \& Denzler \cite{barz2019hierarchy} & 39.73 $\pm$ 0.240 & 6.80 $\pm$ 0.019 & 2.70 $\pm$ 0.022 & 5.48 $\pm$ 0.271 & \cellcolor{highlight} 6.21 $\pm$ 0.005 \\
YOLO-v2 \cite{yolov2} & 33.37 $\pm$ 0.082 & 7.02 $\pm$ 0.004 & 2.34 $\pm$ 0.016 & 5.85 $\pm$ 0.011 & 7.43 $\pm$ 0.016 \\
HXE $\alpha$=0.1 \cite{mbm} & 30.72 $\pm$ 0.036 & 7.00 $\pm$ 0.019 & 2.15 $\pm$ 0.005 & 5.62 $\pm$ 0.008 & 7.08 $\pm$ 0.015\\
HXE $\alpha$=0.6 \cite{mbm} & 34.50 $\pm$ 0.007 & \cellcolor{background} 6.73 $\pm$ 0.014 & 2.32 $\pm$ 0.003 & 5.48 $\pm$ 0.001 & \cellcolor{background} 6.78 $\pm$ 0.003\\
Soft-labels $\beta=30$ \cite{mbm} & 30.53 $\pm$ 0.194 & 7.05 $\pm$ 0.009 & 2.15 $\pm$ 0.013 & 5.66 $\pm$ 0.002 & 7.14 $\pm$ 0.008\\
Soft-labels $\beta=4$ \cite{mbm} & 38.99 $\pm$ 0.105 & \cellcolor{highlight} 6.60 $\pm$ 0.024 & 2.57 $\pm$ 0.004 & \cellcolor{highlight} 5.13 $\pm$ 0.002 & \cellcolor{highlight} 6.21 $\pm$ 0.001 \\
Chang et al. \cite{chang2021your} & 33.46 $\pm$ 0.026 & 6.99 $\pm$ 0.010 & 2.34 $\pm$ 0.006 & 5.75 $\pm$ 0.005 & 7.34 $\pm$ 0.010 \\
HAF \cite{haf} & \cellcolor{background} 30.50 $\pm$ 0.010 & 7.03 $\pm$ 0.024 & \cellcolor{background} 2.14 $\pm$ 0.008 & 5.62 $\pm$ 0.011 & 6.99 $\pm$ 0.009 \\
\cellcolor{best} HiE (Ours) & \cellcolor{highlight} 29.81 $\pm$ 0.081  & 6.95 $\pm$ 0.013 & \cellcolor{highlight} 2.07 $\pm$ 0.014 &  \cellcolor{background} 5.30 $\pm$ 0.001 &  6.86 $\pm$ 0.001  \\
\addlinespace[0.1cm]
\hline
       & \multicolumn{5}{c}{With CRM}                          \\
       \cline{2-6} 
\addlinespace[0.1cm]
Cross-Entropy \cite{crm} & \cellcolor{background} 30.56 $\pm$ 0.020 & \cellcolor{background} 7.01 $\pm$ 0.007 & \cellcolor{background} 2.14 $\pm$ 0.006 & \cellcolor{highlight} 4.93 $\pm$ 0.002 & \cellcolor{highlight} 6.11 $\pm$ 0.001 \\
Cross-Entropy-H & 32.93 $\pm$ 0.029 & 7.05 $\pm$ 0.006 & 2.32 $\pm$ 0.004 & 5.01 $\pm$ 0.008 & 6.14 $\pm$ 0.003 \\
HiE-Self & 30.79 $\pm$ 0.031 & 7.04 $\pm$ 0.008 & 2.17 $\pm$ 0.004 & 4.98 $\pm$ 0.005 & 6.14 $\pm$ 0.002 \\
Barz \& Denzler & 83.55 $\pm$ 0.000 & 11.94 $\pm$ 0.000 & 11.92 $\pm$ 0.000 & 11.91 $\pm$ 0.000 & 11.91 $\pm$ 0.000 \\
YOLO-v2 & 33.98 $\pm$ 0.099 & 6.99 $\pm$ 0.011 & 2.38 $\pm$ 0.012 & 5.05 $\pm$ 0.001 & 6.17 $\pm$ 0.001 \\
HXE $\alpha$=0.1 & 30.80 $\pm$ 0.079 & 6.95 $\pm$ 0.021 & 2.14 $\pm$ 0.005 & \cellcolor{background} 4.94 $\pm$ 0.003 & \cellcolor{highlight} 6.11 $\pm$ 0.002 \\
HXE $\alpha$=0.6 & 34.68 $\pm$ 0.003 & 6.69 $\pm$ 0.007 & 2.32 $\pm$ 0.001 & 4.99 $\pm$ 0.005 & \cellcolor{background} 6.13 $\pm$ 0.003 \\
Soft-labels $\beta=30$ & 30.69 $\pm$ 0.125 & 6.99 $\pm$ 0.007 & 2.15 $\pm$ 0.008 & 4.95 $\pm$ 0.001 & \cellcolor{highlight} 6.11 $\pm$ 0.001 \\
Soft-labels $\beta=4$ & 82.72 $\pm$ 0.079 & 7.54 $\pm$ 0.001 & 6.24 $\pm$ 0.005 & 6.94 $\pm$ 0.005 & 7.25 $\pm$ 0.002 \\
Chang et al. \cite{chang2021your} & 33.73 $\pm$ 0.033 & \cellcolor{highlight} 6.93 $\pm$ 0.015 & 2.34 $\pm$ 0.002 & 5.02 $\pm$ 0.007 & 6.15 $\pm$ 0.001 \\
 HAF \cite{haf} & 30.63 $\pm$ 0.007 & 6.97 $\pm$ 0.024 & \cellcolor{background} 2.14 $\pm$ 0.008 & 4.95 $\pm$ 0.004 & \cellcolor{highlight} 6.11 $\pm$ 0.001 \\
\cellcolor{best} HiE (Ours) & \cellcolor{highlight} 29.89 $\pm$ 0.082 &  \cellcolor{highlight} 6.93 $\pm$ 0.013  & \cellcolor{highlight}  2.07 $\pm$ 0.013 & \cellcolor{highlight} 4.93 $\pm$ 0.001  &  \cellcolor{highlight} 6.11 $\pm$ 0.001 \\
\addlinespace[0.1cm]
\hline
\end{tabular}
\renewcommand{\arraystretch}{1} 
\end{center}
\caption{Results comparing top-$1$ error(\%) and hierarchical metrics on the test set of \textit{tieredImageNet-H}. The \textit{Top} block reports results without using CRM \cite{crm} and the \textit{Bottom} block are reported using CRM. Methods highlighted with {\textcolor{best}{aqua green}} are the best performing methods in top-$1$ error (\%). We use {\textcolor{highlight}{dark green}} for the top performers and {\textcolor{background}{light green}} for the runners-up in each metric.}
\label{tab:tiereimagenet}
\end{table*}

%% file: tables/ssl.tex
\begin{table*}[t]
\centering
\resizebox{\textwidth}{!}{%
\begin{tabular}{@{}cccccccccccccc@{}}
\toprule
\multirow{3}{*}{Methods} &
  \multirow{3}{*}{\#Imgs/Label} &
  \multicolumn{6}{c}{without HiE} &
  \multicolumn{6}{c}{with HiE} \\ \cmidrule(l){3-14} 
 &
   &
  \multicolumn{2}{c}{Top-1 Acc} &
  \multicolumn{2}{c}{Avg. Mistake Severity} &
  \multicolumn{2}{c|}{Hierarchical Distance@1} &
  \multicolumn{2}{c}{Top-1 Acc} &
  \multicolumn{2}{c}{Avg. Mistake Severity} &
  \multicolumn{2}{c}{Hierarchical Distance@1} \\ \cmidrule(l){3-14} 
 &
   &
  w/o CRM &
  w CRM &
  w/o CRM &
  w CRM &
  w/o CRM &
  \multicolumn{1}{c|}{w CRM} &
  w/o CRM &
  w CRM &
  w/o CRM &
  w CRM &
  w/o CRM &
  w CRM \\ \midrule
\multicolumn{1}{c|}{CrossEntropy} &
  \multicolumn{1}{c|}{\multirow{9}{*}{100}} &
  53.90 &
  53.93 &
  2.55 &
  2.48 &
  1.17 &
  \multicolumn{1}{c|}{1.14} &
  56.95 &
  56.82 &
  2.07 &
  2.06 &
  0.89 &
  0.89 \\
\multicolumn{1}{c|}{Pseudo-Label} &
  \multicolumn{1}{c|}{} &
  51.25 &
  51.09 &
  2.52 &
  2.46 &
  1.23 &
  \multicolumn{1}{c|}{1.20} &
  54.78 &
  54.61 &
  2.04 &
  \textbf{2.03} &
  0.92 &
  0.92 \\
\multicolumn{1}{c|}{MoCo} &
  \multicolumn{1}{c|}{} &
  54.72 &
  54.51 &
  2.41 &
  2.36 &
  1.09 &
  \multicolumn{1}{c|}{1.07} &
  57.07 &
  56.91 &
  2.05 &
  2.04 &
  0.88 &
  0.88 \\
\multicolumn{1}{c|}{MocoST} &
  \multicolumn{1}{c|}{} &
  55.77 &
  55.59 &
  2.42 &
  2.37 &
  1.07 &
  \multicolumn{1}{c|}{1.05} &
  57.99 &
  57.78 &
  2.07 &
  2.06 &
  0.87 &
  0.87 \\
\multicolumn{1}{c|}{CrossEntropy-H} &
  \multicolumn{1}{c|}{} &
  55.76 &
  55.71 &
  2.48 &
  2.42 &
  1.10 &
  \multicolumn{1}{c|}{1.07} &
  58.50 &
  58.32 &
  2.08 &
  2.07 &
  \textbf{0.86} &
  \textbf{0.86} \\
\multicolumn{1}{c|}{PseudoLabel-H} &
  \multicolumn{1}{c|}{} &
  56.18 &
  55.98 &
  2.47 &
  2.41 &
  1.08 &
  \multicolumn{1}{c|}{1.06} &
  \textbf{58.80} &
  58.79 &
  2.09 &
  2.08 &
  \textbf{0.86} &
  \textbf{0.86} \\
\multicolumn{1}{c|}{ST-H} &
  \multicolumn{1}{c|}{} &
  54.82 &
  54.71 &
  2.47 &
  2.41 &
  1.12 &
  \multicolumn{1}{c|}{1.09} &
  57.51 &
  57.45 &
  2.07 &
  2.07 &
  0.88 &
  0.88 \\
\multicolumn{1}{c|}{Moco-H} &
  \multicolumn{1}{c|}{} &
  56.05 &
  55.85 &
  2.47 &
  2.42 &
  1.09 &
  \multicolumn{1}{c|}{1.07} &
  58.56 &
  58.49 &
  2.09 &
  2.09 &
  0.87 &
  0.87 \\
\multicolumn{1}{c|}{MocoST-H} &
  \multicolumn{1}{c|}{} &
  55.74 &
  55.49 &
  2.47 &
  2.41 &
  1.10 &
  \multicolumn{1}{c|}{1.07} &
  58.32 &
  58.17 &
  2.08 &
  2.08 &
  0.87 &
  0.87 \\ \midrule
\multicolumn{1}{c|}{CrossEntropy} &
  \multicolumn{1}{c|}{\multirow{9}{*}{50}} &
  45.47 &
  45.60 &
  2.70 &
  2.61 &
  1.47 &
  \multicolumn{1}{c|}{1.42} &
  50.39 &
  50.29 &
  2.00 &
  1.99 &
  0.99 &
  0.99 \\
\multicolumn{1}{c|}{Pseudo-Label} &
  \multicolumn{1}{c|}{} &
  45.97 &
  45.84 &
  2.57 &
  2.51 &
  1.39 &
  \multicolumn{1}{c|}{1.36} &
  50.59 &
  50.43 &
  \textbf{1.99} &
  1.98 &
  0.98 &
  0.98 \\
\multicolumn{1}{c|}{MoCo} &
  \multicolumn{1}{c|}{} &
  47.94 &
  47.75 &
  2.55 &
  2.50 &
  1.33 &
  \multicolumn{1}{c|}{1.31} &
  51.72 &
  51.61 &
  2.01 &
  2.00 &
  0.97 &
  0.97 \\
\multicolumn{1}{c|}{MocoST} &
  \multicolumn{1}{c|}{} &
  49.31 &
  49.20 &
  2.54 &
  2.48 &
  1.29 &
  \multicolumn{1}{c|}{1.26} &
  52.87 &
  52.73 &
  2.02 &
  2.00 &
  0.95 &
  0.95 \\
\multicolumn{1}{c|}{CrossEntropy-H} &
  \multicolumn{1}{c|}{} &
  54.87 &
  54.83 &
  2.50 &
  2.43 &
  1.13 &
  \multicolumn{1}{c|}{1.10} &
  57.69 &
  57.50 &
  2.08 &
  2.07 &
  0.88 &
  0.88 \\
\multicolumn{1}{c|}{PseudoLabel-H} &
  \multicolumn{1}{c|}{} &
  55.17 &
  54.97 &
  2.48 &
  2.40 &
  1.11 &
  \multicolumn{1}{c|}{1.08} &
  \textbf{57.97} &
  57.84 &
  2.07 &
  2.07 &
  \textbf{0.87} &
  \textbf{0.87} \\
\multicolumn{1}{c|}{ST-H} &
  \multicolumn{1}{c|}{} &
  52.42 &
  52.27 &
  2.50 &
  2.43 &
  1.19 &
  \multicolumn{1}{c|}{1.16} &
  55.74 &
  55.65 &
  2.04 &
  2.05 &
  0.90 &
  0.91 \\
\multicolumn{1}{c|}{Moco-H} &
  \multicolumn{1}{c|}{} &
  54.76 &
  54.58 &
  2.48 &
  2.42 &
  1.12 &
  \multicolumn{1}{c|}{1.10} &
  57.65 &
  57.45 &
  2.07 &
  2.08 &
  0.88 &
  0.88 \\
\multicolumn{1}{c|}{MocoST-H} &
  \multicolumn{1}{c|}{} &
  54.53 &
  54.26 &
  2.48 &
  2.42 &
  1.13 &
  \multicolumn{1}{c|}{1.11} &
  57.37 &
  57.21 &
  2.07 &
  2.08 &
  0.88 &
  0.89 \\ \midrule
\multicolumn{1}{c|}{CrossEntropy} &
  \multicolumn{1}{c|}{\multirow{9}{*}{25}} &
  36.47 &
  36.63 &
  2.87 &
  2.77 &
  1.82 &
  \multicolumn{1}{c|}{1.76} &
  43.28 &
  43.22 &
  1.93 &
  \textbf{1.91} &
  1.10 &
  1.09 \\
\multicolumn{1}{c|}{Pseudo-Label} &
  \multicolumn{1}{c|}{} &
  39.13 &
  39.20 &
  2.72 &
  2.66 &
  1.66 &
  \multicolumn{1}{c|}{1.61} &
  44.96 &
  44.91 &
  1.95 &
  1.93 &
  1.07 &
  1.06 \\
\multicolumn{1}{c|}{MoCo} &
  \multicolumn{1}{c|}{} &
  39.48 &
  39.55 &
  2.74 &
  2.68 &
  1.66 &
  \multicolumn{1}{c|}{1.62} &
  45.06 &
  45.00 &
  1.94 &
  1.92 &
  1.06 &
  1.06 \\
\multicolumn{1}{c|}{MocoST} &
  \multicolumn{1}{c|}{} &
  41.22 &
  41.12 &
  2.72 &
  2.63 &
  1.60 &
  \multicolumn{1}{c|}{1.55} &
  46.37 &
  46.30 &
  1.96 &
  1.95 &
  1.05 &
  1.05 \\
\multicolumn{1}{c|}{CrossEntropy-H} &
  \multicolumn{1}{c|}{} &
  53.26 &
  53.18 &
  2.51 &
  2.44 &
  1.17 &
  \multicolumn{1}{c|}{1.14} &
  56.43 &
  56.21 &
  2.06 &
  2.05 &
  0.90 &
  0.90 \\
\multicolumn{1}{c|}{PseudoLabel-H} &
  \multicolumn{1}{c|}{} &
  53.30 &
  53.22 &
  2.51 &
  2.43 &
  1.17 &
  \multicolumn{1}{c|}{1.14} &
  \textbf{56.57} &
  56.44 &
  2.06 &
  2.05 &
  \textbf{0.89} &
  \textbf{0.89} \\
\multicolumn{1}{c|}{ST-H} &
  \multicolumn{1}{c|}{} &
  48.29 &
  48.19 &
  2.59 &
  2.52 &
  1.34 &
  \multicolumn{1}{c|}{1.30} &
  52.69 &
  52.52 &
  2.01 &
  2.01 &
  0.95 &
  0.96 \\
\multicolumn{1}{c|}{Moco-H} &
  \multicolumn{1}{c|}{} &
  53.01 &
  52.97 &
  2.50 &
  2.43 &
  1.17 &
  \multicolumn{1}{c|}{1.14} &
  56.29 &
  56.25 &
  2.06 &
  2.05 &
  0.90 &
  0.90 \\
\multicolumn{1}{c|}{MocoST-H} &
  \multicolumn{1}{c|}{} &
  53.21 &
  53.02 &
  2.49 &
  2.44 &
  1.17 &
  \multicolumn{1}{c|}{1.14} &
  56.34 &
  56.12 &
  2.05 &
  2.06 &
  0.90 &
  0.90 \\ \midrule
\multicolumn{1}{c|}{CrossEntropy} &
  \multicolumn{1}{c|}{\multirow{9}{*}{10}} &
  24.79 &
  25.17 &
  3.12 &
  2.98 &
  2.34 &
  \multicolumn{1}{c|}{2.23} &
  33.33 &
  33.32 &
  1.82 &
  \textbf{1.81} &
  1.22 &
  1.20 \\
\multicolumn{1}{c|}{Pseudo-Label} &
  \multicolumn{1}{c|}{} &
  27.36 &
  27.67 &
  3.01 &
  2.91 &
  2.19 &
  \multicolumn{1}{c|}{2.11} &
  35.08 &
  35.12 &
  1.87 &
  1.86 &
  1.22 &
  1.21 \\
\multicolumn{1}{c|}{MoCo} &
  \multicolumn{1}{c|}{} &
  27.11 &
  26.90 &
  3.01 &
  2.95 &
  2.19 &
  \multicolumn{1}{c|}{2.16} &
  34.95 &
  34.93 &
  1.83 &
  1.82 &
  1.19 &
  1.18 \\
\multicolumn{1}{c|}{MocoST} &
  \multicolumn{1}{c|}{} &
  28.59 &
  28.56 &
  2.95 &
  2.88 &
  2.11 &
  \multicolumn{1}{c|}{2.05} &
  36.03 &
  36.00 &
  1.84 &
  1.82 &
  1.17 &
  1.16 \\
\multicolumn{1}{c|}{CrossEntropy-H} &
  \multicolumn{1}{c|}{} &
  51.36 &
  51.31 &
  2.54 &
  2.47 &
  1.23 &
  \multicolumn{1}{c|}{1.20} &
  54.92 &
  54.80 &
  2.05 &
  2.04 &
  \textbf{0.92} &
  \textbf{0.92} \\
\multicolumn{1}{c|}{PseudoLabel-H} &
  \multicolumn{1}{c|}{} &
  51.35 &
  51.28 &
  2.56 &
  2.49 &
  1.24 &
  \multicolumn{1}{c|}{1.21} &
  55.02 &
  55.02 &
  2.04 &
  2.04 &
  \textbf{0.92} &
  \textbf{0.92} \\
\multicolumn{1}{c|}{ST-H} &
  \multicolumn{1}{c|}{} &
  40.61 &
  40.65 &
  2.77 &
  2.66 &
  1.64 &
  \multicolumn{1}{c|}{1.58} &
  46.67 &
  46.60 &
  1.96 &
  1.94 &
  1.04 &
  1.03 \\
\multicolumn{1}{c|}{Moco-H} &
  \multicolumn{1}{c|}{} &
  51.23 &
  51.27 &
  2.57 &
  2.50 &
  1.25 &
  \multicolumn{1}{c|}{1.22} &
  \textbf{55.13} &
  55.03 &
  2.06 &
  2.06 &
  \textbf{0.92} &
  0.93 \\
\multicolumn{1}{c|}{MocoST-H} &
  \multicolumn{1}{c|}{} &
  50.33 &
  50.26 &
  2.56 &
  2.50 &
  1.27 &
  \multicolumn{1}{c|}{1.24} &
  54.11 &
  54.04 &
  2.04 &
  2.05 &
  0.94 &
  0.94 \\ \bottomrule
\end{tabular}%
}
\caption{Results of the proposed HiE approach on the semi-supervised setting on the \textit{iNaturalist-19} dataset. Best results on each metric are emphasized in \textbf{bold}.}
\label{tab:ssl}
\end{table*}

%% file: tables/architecture_ablation.tex
\begin{table*}[t]
\resizebox{\columnwidth}{!}{%
\renewcommand{\arraystretch}{1.25} 
\begin{tabular}{cccc|ccc|ccc}
\hline
\multirow{2}{*}{Model} & \multicolumn{3}{c|}{Vanilla}                 & \multicolumn{3}{c|}{CRM}                     & \multicolumn{3}{c}{HiE}                      \\ \cline{2-10} 
                       & Top 1 Acc. & Mistake Severity & Hier. Dist@1 & Top 1 Acc. & Mistake Severity & Hier. Dist@1 & Top 1 Acc. & Mistake Severity & Hier. Dist@1 \\ \hline
\multicolumn{1}{c|}{Mobilenet\_V3}   & 45.86 & 2.89 & 1.56 & 45.83 & 2.79 & 1.51 & 46.66 & 2.69 & 1.44 \\
\multicolumn{1}{c|}{ResNet18}        & 63.63 & 2.39 & 0.87 & 63.65 & 2.31 & 0.84 & 64.65 & 2.15 & 0.76 \\
\multicolumn{1}{c|}{ResNet50}        & 69.48 & 2.23 & 0.68 & 69.49 & 2.19 & 0.67 & 70.34 & 2.03 & 0.60 \\
\multicolumn{1}{c|}{ResNet101}       & 70.86 & 2.14 & 0.62 & 70.87 & 2.11 & 0.61 & 71.66 & 1.96 & 0.56 \\
\multicolumn{1}{c|}{EfficientNet-B0} & 67.69 & 2.25 & 0.73 & 67.68 & 2.21 & 0.71 & 68.75 & 2.01 & 0.63 \\
\multicolumn{1}{c|}{DenseNet121}     & 67.86 & 2.27 & 0.73 & 67.89 & 2.23 & 0.72 & 68.97 & 2.04 & 0.63 \\
\multicolumn{1}{c|}{DeiT}            & 64.71 & 2.38 & 0.84 & 64.74 & 2.32 & 0.82 & 65.73 & 2.14 & 0.73 \\
\multicolumn{1}{c|}{ViT}             & 66.19 & 2.36 & 0.80 & 66.27 & 2.29 & 0.77 & 67.56 & 2.06 & 0.67 \\
\multicolumn{1}{c|}{SwinT}           & 75.08 & 2.10 & 0.52 & 75.09 & 2.06 & 0.51 & 75.74 & 1.88 & 0.46 \\ \hline
\end{tabular}%
}
\caption{Results of the proposed HiE approach with different architectures on \textit{species} taxonomy of the \textit{iNaturalist-19} dataset.}
\label{tab:arch_ablation}
\renewcommand{\arraystretch}{1} 
\end{table*}

%% file: tables/multi_level.tex
\begin{table*}[t!]
\centering
\resizebox{0.8\columnwidth}{!}{%
\begin{tabular}{@{}cccccc@{}}
\toprule
Hier. Levels  & Top-1 Acc.     & Mistake Severity  & Hier. Dist @ 1 & Hier. Dist @ 5 & Hier. Dist @ 20 \\ \midrule
\{7\}      & 63.63          & 2.39              & 0.87           & 1.97           & 3.25            \\
\{6, 7\}   & 64.65          & 2.15              & 0.76           & 1.33           & 2.19            \\
\{5 - 7\}  & \textbf{64.72} & \textbf{2.09}     & \textbf{0.74}  & \textbf{1.26}  & 2.00            \\
\{4 - 7\}  & 64.48          & 2.12              & 0.75           & \textbf{1.26}  & 1.91   \\ 
\{3 - 7\}  & 64.47          & 2.12              & 0.75           & \textbf{1.26}  & \textbf{1.89}   \\
\{2 - 7\}  & 64.46          & 2.13              & 0.76           & 1.27           & \textbf{1.89}   \\
\{1 - 7\}  & 64.44          & 2.14              & 0.76           & 1.27           & \textbf{1.89}   \\ \bottomrule
\end{tabular}
}
\caption{Results of the proposed HiE approach on fine-grained level (\textit{species}) of \textit{iNaturalist-19} dataset when incorporating multiple hierarchical levels. Best results are highlighted in \textbf{bold}.}
\label{tab:multi_level}
\end{table*}

%% file: neurips_2023.bbl
\begin{thebibliography}{50}
\providecommand{\natexlab}[1]{#1}
\providecommand{\url}[1]{\texttt{#1}}
\expandafter\ifx\csname urlstyle\endcsname\relax
  \providecommand{\doi}[1]{doi: #1}\else
  \providecommand{\doi}{doi: \begingroup \urlstyle{rm}\Url}\fi

\bibitem[Angelova and Zhu(2013)]{angelova2013efficient}
Anelia Angelova and Shenghuo Zhu.
\newblock Efficient object detection and segmentation for fine-grained
  recognition.
\newblock In \emph{Proceedings of the IEEE conference on computer vision and
  pattern recognition}, pages 811--818, 2013.

\bibitem[Barz and Denzler(2019)]{barz2019hierarchy}
Bj{\"o}rn Barz and Joachim Denzler.
\newblock Hierarchy-based image embeddings for semantic image retrieval.
\newblock In \emph{WACV}, 2019.

\bibitem[Bertinetto et~al.(2020)Bertinetto, Mueller, Tertikas, Samangooei, and
  Lord]{mbm}
Luca Bertinetto, Romain Mueller, Konstantinos Tertikas, Sina Samangooei, and
  Nicholas~A Lord.
\newblock Making better mistakes: Leveraging class hierarchies with deep
  networks.
\newblock In \emph{Proceedings of the IEEE/CVF Conference on Computer Vision
  and Pattern Recognition}, pages 12506--12515, 2020.

\bibitem[Bilal et~al.(2017)Bilal, Jourabloo, Ye, Liu, and
  Ren]{bilal2017convolutional}
Alsallakh Bilal, Amin Jourabloo, Mao Ye, Xiaoming Liu, and Liu Ren.
\newblock Do convolutional neural networks learn class hierarchy?
\newblock \emph{IEEE transactions on visualization and computer graphics},
  24\penalty0 (1):\penalty0 152--162, 2017.

\bibitem[Brown et~al.(2020)Brown, Mann, Ryder, Subbiah, Kaplan, Dhariwal,
  Neelakantan, Shyam, Sastry, Askell, et~al.]{brown2020language}
Tom Brown, Benjamin Mann, Nick Ryder, Melanie Subbiah, Jared~D Kaplan, Prafulla
  Dhariwal, Arvind Neelakantan, Pranav Shyam, Girish Sastry, Amanda Askell,
  et~al.
\newblock Language models are few-shot learners.
\newblock \emph{NeurIPS}, 2020.

\bibitem[Chang et~al.(2021)Chang, Pang, Zheng, Ma, Song, and
  Guo]{chang2021your}
Dongliang Chang, Kaiyue Pang, Yixiao Zheng, Zhanyu Ma, Yi-Zhe Song, and Jun
  Guo.
\newblock Your" flamingo" is my" bird": Fine-grained, or not.
\newblock In \emph{Proceedings of the IEEE/CVF Conference on Computer Vision
  and Pattern Recognition}, pages 11476--11485, 2021.

\bibitem[Deng et~al.(2010)Deng, Berg, Li, and Fei-Fei]{deng2010does}
Jia Deng, Alexander~C Berg, Kai Li, and Li~Fei-Fei.
\newblock What does classifying more than 10,000 image categories tell us?
\newblock In \emph{European conference on computer vision}, pages 71--84.
  Springer, 2010.

\bibitem[Domingos(1999)]{domingos1999metacost}
Pedro Domingos.
\newblock Metacost: A general method for making classifiers cost-sensitive.
\newblock In \emph{KDD}, 1999.

\bibitem[Dosovitskiy et~al.(2021)Dosovitskiy, Beyer, Kolesnikov, Weissenborn,
  Zhai, Unterthiner, Dehghani, Minderer, Heigold, Gelly, Uszkoreit, and
  Houlsby]{dosovitskiy2020vit}
Alexey Dosovitskiy, Lucas Beyer, Alexander Kolesnikov, Dirk Weissenborn,
  Xiaohua Zhai, Thomas Unterthiner, Mostafa Dehghani, Matthias Minderer, Georg
  Heigold, Sylvain Gelly, Jakob Uszkoreit, and Neil Houlsby.
\newblock An image is worth 16x16 words: Transformers for image recognition at
  scale.
\newblock \emph{ICLR}, 2021.

\bibitem[Duda et~al.(1973)Duda, Hart, et~al.]{duda1973pattern}
Richard~O Duda, Peter~E Hart, et~al.
\newblock \emph{Pattern classification and scene analysis}, volume~3.
\newblock Wiley New York, 1973.

\bibitem[Frome et~al.(2013)Frome, Corrado, Shlens, Bengio, Dean, Ranzato, and
  Mikolov]{devise}
Andrea Frome, Greg~S Corrado, Jon Shlens, Samy Bengio, Jeff Dean, Marc'Aurelio
  Ranzato, and Tomas Mikolov.
\newblock Devise: A deep visual-semantic embedding model.
\newblock \emph{Advances in neural information processing systems}, 26, 2013.

\bibitem[Garg et~al.(2022)Garg, Sani, and Anand]{haf}
Ashima Garg, Depanshu Sani, and Saket Anand.
\newblock Learning hierarchy aware features for reducing mistake severity.
\newblock In \emph{European Conference on Computer Vision}, pages 252--267.
  Springer, 2022.

\bibitem[Garnot and Landrieu(2021)]{garnot2020leveraging}
Vivien Sainte~Fare Garnot and Loic Landrieu.
\newblock Leveraging class hierarchies with metric-guided prototype learning.
\newblock In \emph{BMVC}, 2021.

\bibitem[He et~al.(2016)He, Zhang, Ren, and Sun]{he2016deep}
Kaiming He, Xiangyu Zhang, Shaoqing Ren, and Jian Sun.
\newblock Deep residual learning for image recognition.
\newblock In \emph{Proceedings of the IEEE conference on computer vision and
  pattern recognition}, pages 770--778, 2016.

\bibitem[He et~al.(2020)He, Fan, Wu, Xie, and Girshick]{moco}
Kaiming He, Haoqi Fan, Yuxin Wu, Saining Xie, and Ross Girshick.
\newblock Momentum contrast for unsupervised visual representation learning.
\newblock In \emph{Proceedings of the IEEE/CVF conference on computer vision
  and pattern recognition}, pages 9729--9738, 2020.

\bibitem[Hinton et~al.(2015)Hinton, Vinyals, Dean,
  et~al.]{hinton2015distilling}
Geoffrey Hinton, Oriol Vinyals, Jeff Dean, et~al.
\newblock Distilling the knowledge in a neural network.
\newblock \emph{arXiv preprint arXiv:1503.02531}, 2\penalty0 (7), 2015.

\bibitem[Howard et~al.(2017)Howard, Zhu, Chen, Kalenichenko, Wang, Weyand,
  Andreetto, and Adam]{howard2017mobilenets}
Andrew~G Howard, Menglong Zhu, Bo~Chen, Dmitry Kalenichenko, Weijun Wang,
  Tobias Weyand, Marco Andreetto, and Hartwig Adam.
\newblock Mobilenets: Efficient convolutional neural networks for mobile vision
  applications.
\newblock \emph{arXiv preprint arXiv:1704.04861}, 2017.

\bibitem[Huang et~al.(2017)Huang, Liu, Van Der~Maaten, and
  Weinberger]{huang2017densely}
Gao Huang, Zhuang Liu, Laurens Van Der~Maaten, and Kilian~Q Weinberger.
\newblock Densely connected convolutional networks.
\newblock In \emph{Proceedings of the IEEE conference on computer vision and
  pattern recognition}, pages 4700--4708, 2017.

\bibitem[Karthik et~al.(2021)Karthik, Prabhu, Dokania, and Gandhi]{crm}
Shyamgopal Karthik, Ameya Prabhu, Puneet~K Dokania, and Vineet Gandhi.
\newblock No cost likelihood manipulation at test time for making better
  mistakes in deep networks.
\newblock \emph{ICLR}, 2021.

\bibitem[Khrulkov et~al.(2020)Khrulkov, Mirvakhabova, Ustinova, Oseledets, and
  Lempitsky]{khrulkov2020hyperbolic}
Valentin Khrulkov, Leyla Mirvakhabova, Evgeniya Ustinova, Ivan Oseledets, and
  Victor Lempitsky.
\newblock Hyperbolic image embeddings.
\newblock In \emph{Proceedings of the IEEE/CVF Conference on Computer Vision
  and Pattern Recognition}, pages 6418--6428, 2020.

\bibitem[Lee et~al.(2013)]{pseudo}
Dong-Hyun Lee et~al.
\newblock Pseudo-label: The simple and efficient semi-supervised learning
  method for deep neural networks.
\newblock In \emph{Workshop on challenges in representation learning, ICML},
  2013.

\bibitem[Lin et~al.(2015)Lin, Shen, Lu, and Jia]{lin2015deep}
Di~Lin, Xiaoyong Shen, Cewu Lu, and Jiaya Jia.
\newblock Deep lac: Deep localization, alignment and classification for
  fine-grained recognition.
\newblock In \emph{Proceedings of the IEEE conference on computer vision and
  pattern recognition}, pages 1666--1674, 2015.

\bibitem[Liu et~al.(2020)Liu, Chen, Pan, Ngo, Chua, and
  Jiang]{liu2020hyperbolic}
Shaoteng Liu, Jingjing Chen, Liangming Pan, Chong-Wah Ngo, Tat-Seng Chua, and
  Yu-Gang Jiang.
\newblock Hyperbolic visual embedding learning for zero-shot recognition.
\newblock In \emph{Proceedings of the IEEE/CVF Conference on Computer Vision
  and Pattern Recognition}, pages 9273--9281, 2020.

\bibitem[Liu et~al.(2021)Liu, Lin, Cao, Hu, Wei, Zhang, Lin, and
  Guo]{liu2021Swin}
Ze~Liu, Yutong Lin, Yue Cao, Han Hu, Yixuan Wei, Zheng Zhang, Stephen Lin, and
  Baining Guo.
\newblock Swin transformer: Hierarchical vision transformer using shifted
  windows.
\newblock In \emph{Proceedings of the IEEE/CVF International Conference on
  Computer Vision (ICCV)}, 2021.

\bibitem[Maji et~al.(2013)Maji, Rahtu, Kannala, Blaschko, and
  Vedaldi]{aircraft}
Subhransu Maji, Esa Rahtu, Juho Kannala, Matthew Blaschko, and Andrea Vedaldi.
\newblock Fine-grained visual classification of aircraft.
\newblock \emph{arXiv preprint arXiv:1306.5151}, 2013.

\bibitem[Miller(1995)]{wordnet}
George~A Miller.
\newblock Wordnet: a lexical database for english.
\newblock \emph{Communications of the ACM}, 38\penalty0 (11):\penalty0 39--41,
  1995.

\bibitem[Nickel and Kiela(2017)]{nickel2017poincare}
Maximillian Nickel and Douwe Kiela.
\newblock Poincar{\'e} embeddings for learning hierarchical representations.
\newblock In \emph{Advances in neural information processing systems}, pages
  6338--6347, 2017.

\bibitem[Novack et~al.(2023)Novack, McAuley, Lipton, and Garg]{chils}
Zachary Novack, Julian McAuley, Zachary~Chase Lipton, and Saurabh Garg.
\newblock Chils: Zero-shot image classification with hierarchical label sets.
\newblock In \emph{ICML}, 2023.

\bibitem[Phoo and Hariharan(2021)]{phoo2021coarsely}
Cheng~Perng Phoo and Bharath Hariharan.
\newblock Coarsely-labeled data for better few-shot transfer.
\newblock In \emph{Proceedings of the IEEE/CVF International Conference on
  Computer Vision}, pages 9052--9061, 2021.

\bibitem[Radford et~al.(2021)Radford, Kim, Hallacy, Ramesh, Goh, Agarwal,
  Sastry, Askell, Mishkin, Clark, et~al.]{clip}
Alec Radford, Jong~Wook Kim, Chris Hallacy, Aditya Ramesh, Gabriel Goh,
  Sandhini Agarwal, Girish Sastry, Amanda Askell, Pamela Mishkin, Jack Clark,
  et~al.
\newblock Learning transferable visual models from natural language
  supervision.
\newblock pages 8748--8763, 2021.

\bibitem[Redmon and Farhadi(2017)]{yolov2}
Joseph Redmon and Ali Farhadi.
\newblock Yolo9000: better, faster, stronger.
\newblock In \emph{Proceedings of the IEEE conference on computer vision and
  pattern recognition}, pages 7263--7271, 2017.

\bibitem[Ren et~al.(2018)Ren, Triantafillou, Ravi, Snell, Swersky, Tenenbaum,
  Larochelle, and Zemel]{ren2018meta}
Mengye Ren, Eleni Triantafillou, Sachin Ravi, Jake Snell, Kevin Swersky,
  Joshua~B Tenenbaum, Hugo Larochelle, and Richard~S Zemel.
\newblock Meta-learning for semi-supervised few-shot classification.
\newblock \emph{arXiv preprint arXiv:1803.00676}, 2018.

\bibitem[Ridnik et~al.(2021)Ridnik, Ben-Baruch, Noy, and
  Zelnik-Manor]{ridnik2021imagenet}
Tal Ridnik, Emanuel Ben-Baruch, Asaf Noy, and Lihi Zelnik-Manor.
\newblock Imagenet-21k pretraining for the masses.
\newblock \emph{arXiv preprint arXiv:2104.10972}, 2021.

\bibitem[Rizve et~al.(2021)Rizve, Duarte, Rawat, and Shah]{pseudo++}
Mamshad~Nayeem Rizve, Kevin Duarte, Yogesh~S Rawat, and Mubarak Shah.
\newblock In defense of pseudo-labeling: An uncertainty-aware pseudo-label
  selection framework for semi-supervised learning.
\newblock \emph{arXiv preprint arXiv:2101.06329}, 2021.

\bibitem[Silla and Freitas(2011)]{silla2011survey}
Carlos~N Silla and Alex~A Freitas.
\newblock A survey of hierarchical classification across different application
  domains.
\newblock \emph{Data Mining and Knowledge Discovery}, 22\penalty0 (1):\penalty0
  31--72, 2011.

\bibitem[Sohn et~al.(2020)Sohn, Berthelot, Carlini, Zhang, Zhang, Raffel,
  Cubuk, Kurakin, and Li]{fixmatch}
Kihyuk Sohn, David Berthelot, Nicholas Carlini, Zizhao Zhang, Han Zhang,
  Colin~A Raffel, Ekin~Dogus Cubuk, Alexey Kurakin, and Chun-Liang Li.
\newblock Fixmatch: Simplifying semi-supervised learning with consistency and
  confidence.
\newblock \emph{Advances in neural information processing systems},
  33:\penalty0 596--608, 2020.

\bibitem[Su and Maji(2021)]{su2021semi}
Jong-Chyi Su and Subhransu Maji.
\newblock Semi-supervised learning with taxonomic labels.
\newblock \emph{arXiv preprint arXiv:2111.11595}, 2021.

\bibitem[Su et~al.(2021)Su, Cheng, and Maji]{semi-inat}
Jong-Chyi Su, Zezhou Cheng, and Subhransu Maji.
\newblock A realistic evaluation of semi-supervised learning for fine-grained
  classification.
\newblock In \emph{Proceedings of the IEEE/CVF Conference on Computer Vision
  and Pattern Recognition}, pages 12966--12975, 2021.

\bibitem[Tan and Le(2019)]{tan2019efficientnet}
Mingxing Tan and Quoc Le.
\newblock Efficientnet: Rethinking model scaling for convolutional neural
  networks.
\newblock In \emph{International conference on machine learning}, pages
  6105--6114. PMLR, 2019.

\bibitem[Touvron et~al.(2021)Touvron, Cord, Douze, Massa, Sablayrolles, and
  J{\'e}gou]{touvron2021training}
Hugo Touvron, Matthieu Cord, Matthijs Douze, Francisco Massa, Alexandre
  Sablayrolles, and Herv{\'e} J{\'e}gou.
\newblock Training data-efficient image transformers \& distillation through
  attention.
\newblock In \emph{International conference on machine learning}, pages
  10347--10357. PMLR, 2021.

\bibitem[Valmadre(2022)]{valmadre2022hierarchical}
Jack Valmadre.
\newblock Hierarchical classification at multiple operating points.
\newblock In \emph{NeurIPS}, 2022.

\bibitem[Van~Horn et~al.(2018{\natexlab{a}})Van~Horn, Mac~Aodha, Song, Cui,
  Sun, Shepard, Adam, Perona, and Belongie]{inat}
Grant Van~Horn, Oisin Mac~Aodha, Yang Song, Yin Cui, Chen Sun, Alex Shepard,
  Hartwig Adam, Pietro Perona, and Serge Belongie.
\newblock The inaturalist species classification and detection dataset.
\newblock In \emph{Proceedings of the IEEE conference on computer vision and
  pattern recognition}, pages 8769--8778, 2018{\natexlab{a}}.

\bibitem[Van~Horn et~al.(2018{\natexlab{b}})Van~Horn, Mac~Aodha, Song, Cui,
  Sun, Shepard, Adam, Perona, and Belongie]{van2018inaturalist}
Grant Van~Horn, Oisin Mac~Aodha, Yang Song, Yin Cui, Chen Sun, Alex Shepard,
  Hartwig Adam, Pietro Perona, and Serge Belongie.
\newblock The inaturalist species classification and detection dataset.
\newblock In \emph{Proceedings of the IEEE conference on computer vision and
  pattern recognition}, pages 8769--8778, 2018{\natexlab{b}}.

\bibitem[Wah et~al.(2011)Wah, Branson, Welinder, Perona, and Belongie]{birds}
Catherine Wah, Steve Branson, Peter Welinder, Pietro Perona, and Serge
  Belongie.
\newblock The caltech-ucsd birds-200-2011 dataset.
\newblock 2011.

\bibitem[Wang et~al.(2020)Wang, Kondratyuk, Christiansen, Kitani, Alon, and
  Eban]{wang2020wisdom}
Xiaofang Wang, Dan Kondratyuk, Eric Christiansen, Kris~M Kitani, Yair Alon, and
  Elad Eban.
\newblock Wisdom of committees: An overlooked approach to faster and more
  accurate models.
\newblock \emph{arXiv preprint arXiv:2012.01988}, 2020.

\bibitem[Wu et~al.(2016)Wu, Merler, Uceda-Sosa, and Smith]{wu2016learning}
Hui Wu, Michele Merler, Rosario Uceda-Sosa, and John~R Smith.
\newblock Learning to make better mistakes: Semantics-aware visual food
  recognition.
\newblock In \emph{Proceedings of the 24th ACM international conference on
  Multimedia}, pages 172--176, 2016.

\bibitem[Xian et~al.(2016)Xian, Akata, Sharma, Nguyen, Hein, and
  Schiele]{latem}
Yongqin Xian, Zeynep Akata, Gaurav Sharma, Quynh Nguyen, Matthias Hein, and
  Bernt Schiele.
\newblock Latent embeddings for zero-shot classification.
\newblock In \emph{Proceedings of the IEEE conference on computer vision and
  pattern recognition}, pages 69--77, 2016.

\bibitem[Xian et~al.(2018)Xian, Lampert, Schiele, and Akata]{xian2018zero}
Yongqin Xian, Christoph~H Lampert, Bernt Schiele, and Zeynep Akata.
\newblock Zero-shot learning—a comprehensive evaluation of the good, the bad
  and the ugly.
\newblock \emph{IEEE transactions on pattern analysis and machine
  intelligence}, 41\penalty0 (9):\penalty0 2251--2265, 2018.

\bibitem[Zadrozny and Elkan(2001)]{zadrozny2001learning}
Bianca Zadrozny and Charles Elkan.
\newblock Learning and making decisions when costs and probabilities are both
  unknown.
\newblock In \emph{KDD}, 2001.

\bibitem[Zhang et~al.(2014)Zhang, Donahue, Girshick, and
  Darrell]{zhang2014part}
Ning Zhang, Jeff Donahue, Ross Girshick, and Trevor Darrell.
\newblock Part-based r-cnns for fine-grained category detection.
\newblock In \emph{European conference on computer vision}, pages 834--849.
  Springer, 2014.

\end{thebibliography}
